\documentclass[letterpaper, 10 pt, conference]{ieeeconf}
\IEEEoverridecommandlockouts
\usepackage[utf8]{inputenc}
\usepackage{graphicx}
\usepackage{multicol}
\usepackage{multirow}
\usepackage{array}
\usepackage{footmisc}
\usepackage{amssymb}
\usepackage{amsmath}
\usepackage{amsfonts}
\usepackage{algorithm}
\usepackage{algpseudocode}
\usepackage{tikz}
\usepackage{float}
\usetikzlibrary{math,shapes}
\usepackage{tabularx}
\usepackage{pbox}
\usepackage{ragged2e}
\usepackage{multirow}
\usepackage{tabularx}
\usepackage{balance}
\usepackage{subfig}
\usepackage{relsize}
\usepackage{hyperref}

\usepackage{listings}
\mathchardef\mhyphen="2D

\newcommand{\squeeze}{\vspace{-0.65cm}}

\usepackage{etoolbox}
\makeatletter
\patchcmd{\@makecaption}
  {\scshape}
  {}
  {}
  {}
\makeatletter
\patchcmd{\@makecaption}
  {\\}
  { -\ }
  {}
  {}
\makeatother
\usepackage{siunitx}
\title{\LARGE \bf Can Machines Garden? Systematically Comparing \\
       the AlphaGarden vs. Professional Horticulturalists}
\author{Simeon Adebola$^{*1}$, Rishi Parikh$^{*1}$, Mark Presten$^{1}$, Satvik Sharma$^{1}$, Shrey Aeron$^{1}$, Ananth \\Rao$^{1}$,  Sandeep Mukherjee$^{1}$, Tomson Qu$^{1}$, Christina Wistrom$^{2}$, Eugen Solowjow$^{3}$ , Ken Goldberg$^{1}$
\thanks{$^{*}$Equal Contribution}
\thanks{$^{1}$The AUTOLab at UC Berkeley (automation.berkeley.edu) {\tt\footnotesize \{simeon.adebola, goldberg\}@berkeley.edu}}%
\thanks{$^{2}$Berkeley GreenHouse {\tt\footnotesize \{cwistrom@berkeley.edu\}}}%
\thanks{$^{3}$Siemens Research Lab, Berkeley, CA {\tt\footnotesize \{eugen.solowjow@siemens.com\}}}%
}

\begin{document}

\maketitle
\thispagestyle{empty}
\pagestyle{empty}

\begin{abstract}
The AlphaGarden is an automated testbed for indoor polyculture farming which combines a first-order plant simulator, a gantry robot, a seed planting algorithm, plant phenotyping and tracking algorithms, irrigation sensors and algorithms, and custom pruning tools and algorithms. In this paper, we systematically compare the performance of the AlphaGarden to professional horticulturalists on the staff of the UC Berkeley Oxford Tract Greenhouse. The humans and the machine tend side-by-side polyculture gardens with the same seed arrangement. We compare performance in terms of canopy coverage, plant diversity, and water consumption. Results from two 60-day cycles suggest that the automated AlphaGarden performs comparably to professional horticulturalists in terms of coverage and diversity, and reduces water consumption by as much as 44\%.

Code, videos, and datasets are available at \href{https://sites.google.com/berkeley.edu/systematiccomparison}{https://sites.google.com/berkeley.edu/systematiccomparison}.

\end{abstract}

\section{Introduction}
In 1950, Alan Turing considered the question “Can Machines Think?" and proposed a test based on comparing human vs. machine ability to answer questions.

In this paper, we consider the question “Can Machines Garden?" based on comparing human vs. machine ability to tend a real polyculture garden. %
Precision polyculture agriculture can reduce pesticides and water usage but requires more labor than monoculture farming \cite{avigalsimulatingtase}. Could robots help?

A robot could change the way agriculture is practiced by increasing yield and improving resource efficiency. Additionally, polyculture gardens more closely resemble nature, reducing the depletion of soil nutrients while exhibiting many benefits compared to monoculture gardens, such as pest reduction, reduced demand for fertilizers, soil support, and crop support  ~\cite{BRACKEN2019483,liebman_polyculture_1987}. 

The AlphaGarden combines AlphaGardenSim, a first-order polyculture simulator, with an open source, commercially available gantry robot ~\cite{Farmbot}. This paper builds on previous work ~\cite{avigal2020simulating,avigallearning,avigalsimulatingtase,presten_automated_2022} by comparing gardens tended by AlphaGarden with ones tended by expert professional horticulturalists, evaluating on coverage, diversity, and water usage. 

\begin{figure}
    \vspace{0.2cm}
    \centering
    \includegraphics[width=85.5mm]{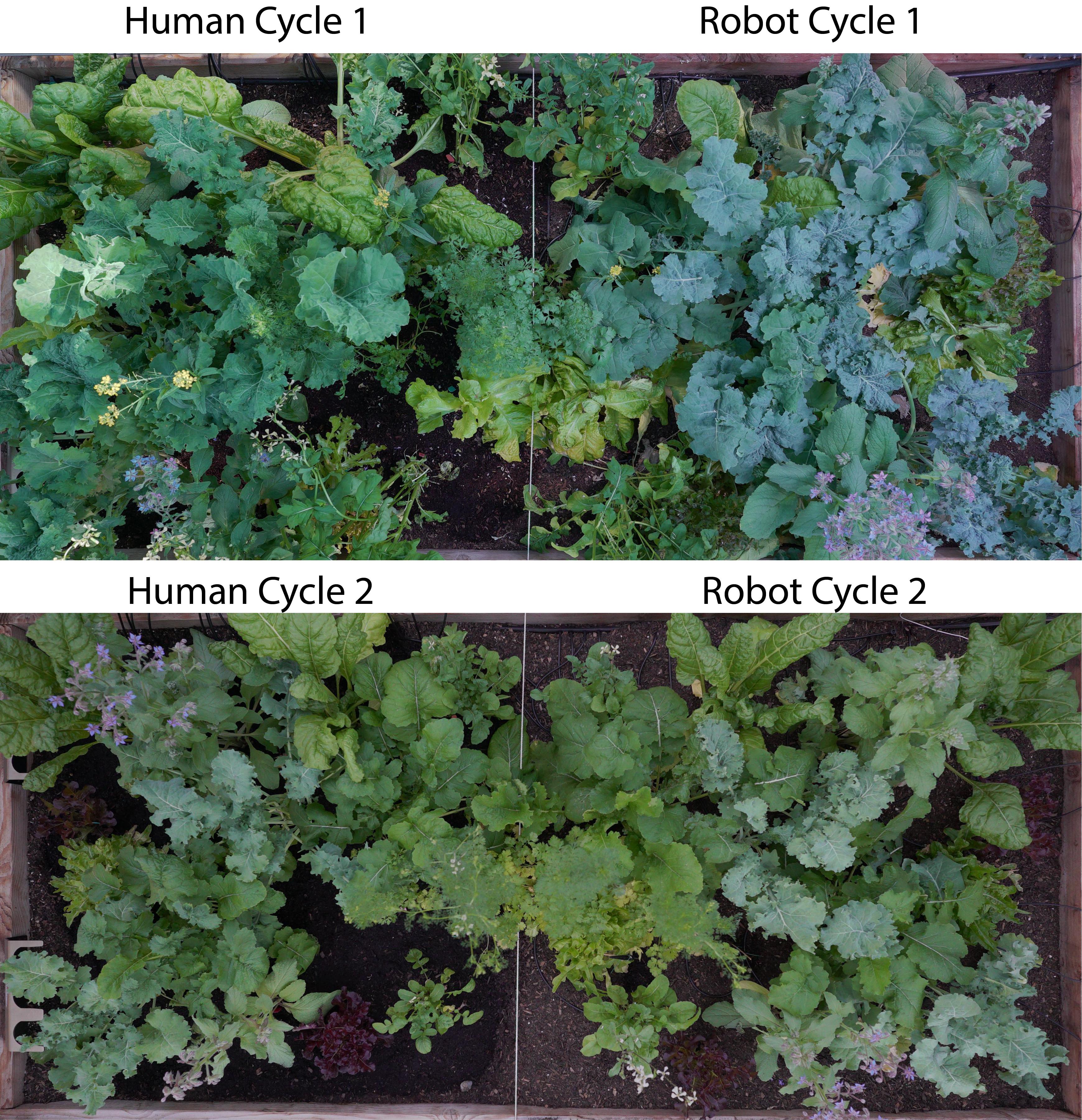}
    \caption{\textbf{Four gardens at day 60.}
    A side-by-side comparison of four 1.5m by 1.5m real gardens planted with mirrored identical seed arrangements (mirrored across the white string in the middle). In both 60-day cycles, the left half was tended by human experts, while the right half was tended by the AlphaGarden robot system. Coverage and Diversity on Day 60 are comparable. The AlphaGarden consumed 37\% and 44\% less water, respectively.
    }
    \vspace{-0.6cm}
\label{fig:cycle_at_60}
\end{figure}

This paper makes three contributions:
\begin{enumerate}
    \item The AlphaGarden system with Integration of sensing and control based on a high-resolution camera and soil moisture sensors, custom simulation, algorithms for monitoring plant growth, and computing prune points. 

   \item Staggered seed planting based on learned plant growth models, and variable and closed-loop drip irrigation
    \item Detailed time-lapse data from 120 days of physical experiments (2 growth cycles) and analysis suggesting the AlphaGarden performs comparably to human horticulturalists in terms of coverage and diversity while reducing water consumption. 
\end{enumerate}

\section{Related Work}

Indoor farming is a class of Urban Agriculture where crops can be grown inside a building ~\cite{https://doi.org/10.1029/2022EF002748}, ~\cite{dorr_environmental_2021}. Recent surveys have studied yields and environmental impacts of indoor farming~\cite{https://doi.org/10.1029/2022EF002748}, ~\cite{dorr_environmental_2021}. There is also increasing commercial interest in indoor farming ~\cite{verbeek_consumer_2022, noauthor_home_nodate_Bowery, noauthor_home_nodate_IronOx}. %

Modelling and simulation can allow us to simulate plant growth and test many potentially useful policies at a much faster rate than natural growth. While monoculture plant simulators are well known ~\cite{stomph2020designing}~\cite{jones2003dssat}~\cite{steduto2009aquacrop}, in prior work we presented the AlphaGardenSim~\cite{avigal2020simulating}, a polyculture plant simulator that uses first-order models of single plant growth, simulating inter-plant dynamics and competition for water and light. The model parameters used in AlphaGardenSim were tuned using real-world measurements from a physical garden testbed. AlphaGardenSim allows us to simulate plant growth and apply automated irrigation and pruning policies to physical environments.

Prior work has used the Farmbot, a robot gantry system, to interface with humans to help with seed planting, watering, and plant monitoring routines~\cite{farmbot2}. In addition, \cite{presten_automated_2022} used the Farmbot to autonomously prune polyculture plants with a pruning pipeline consisting of a camera over the testbed, a Plant Phenotyping Network and a Bounding Disk Tracking algorithm which determine the garden’s state, AlphaGardenSim which decides what plants to prune, a Prune Point Identification network for which leaves to prune, visual servoing by the Farmbot guided by another camera connected to it to arrive at selected leaves, and custom pruning tools. Besides the Farmbot, other robotic systems have also tended gardens such as work from Correll et al. \cite{correll2009building}, who designed a distributed autonomous gardening system with mobile manipulators that detect plants, irrigate, and grasp fruit. Botterill et al. studied pruning of grape vines using a six degree of freedom robot arm \cite{botterill}. Also, Hernandez et al.~\cite{hernandez2019autonomous} developed an autonomous urban garden that monitors soil moisture, temperature, and humidity. Some previous work has also compared the performance of a robotic system designed for agriculture to that of a human. Hayashi et al. developed a strawberry harvesting robot and indicated that its execution time, while better than previous studies~\cite{VANHENTEN2003305, hayashi2003robotic}, took 2.5-3 times longer to harvest than a human \cite{hayashi_evaluation_2010}

Corbett-Davies et al. leveraged classification and search algorithms to make pruning decisions on simulated vines and indicated their system outperformed a human pruner with a skill level above the typical vineyard worker on this simulated domain \cite{corbettdavisautovine}. However, to the best of our knowledge, there is no systematic comparison between an autonomous robotic system and horticulturalist(s) tending two identically planted gardens.

The current state of polyculture agriculture implements several strategies to promote more natural and optimal plant growth such as relay planting and intercropping, which is planting the next cycle of crops before the end of the first cycle and adding plants in empty spaces of the garden \cite{2021}. In AlphaGardenSim, we build on this by dynamically predicting vacancies in the garden created either through pruning or decay of plants and determining which seed to plant to promote coverage and diversity. 

While in traditional farming, both rainfed and irrigated agriculture is discussed~\cite{noauthor_state_2020}, in indoor farming, most of the focus has to be on irrigated agriculture. Moreover, in the face of drought in parts around the world, irrigated agriculture is still the world's major water user, using more than 70 percent of global water~\cite{w11091758, noauthor_state_2020}. Therefore, there is research into sustainable irrigation for agriculture~\cite{w11091758}. In AlphaGardenSim, irrigation simulation is also being improved to further optimize coverage and diversity while reducing water usage. We present some of our ongoing work with irrigation in this paper. 

 Closed-loop robotic irrigation allows for better water management. There exists prior work using mobile robots for irrigation. For example, Zhen et al. prototyped a wireless network for closed-loop drip irrigation that used soil sensors as input \cite{article}. We apply closed-loop drip irrigation in a polyculture setting by interfacing a custom Arduino board (Farmduino \cite{Farmbot}) on the Farmbot with soil moisture sensors through a wireless server.

\section{Problem Statement}
In each garden cycle, several plant types are planted over a specified number of days with the goal of increasing coverage, plant diversity, and reducing water usage. The problem statement below is consistent with \cite{presten_automated_2022}, but adds water usage measurements and comparison with human experts.

\begin{figure}
    \vspace{0.2cm}
    \centering
    \includegraphics[width=88mm, clip]{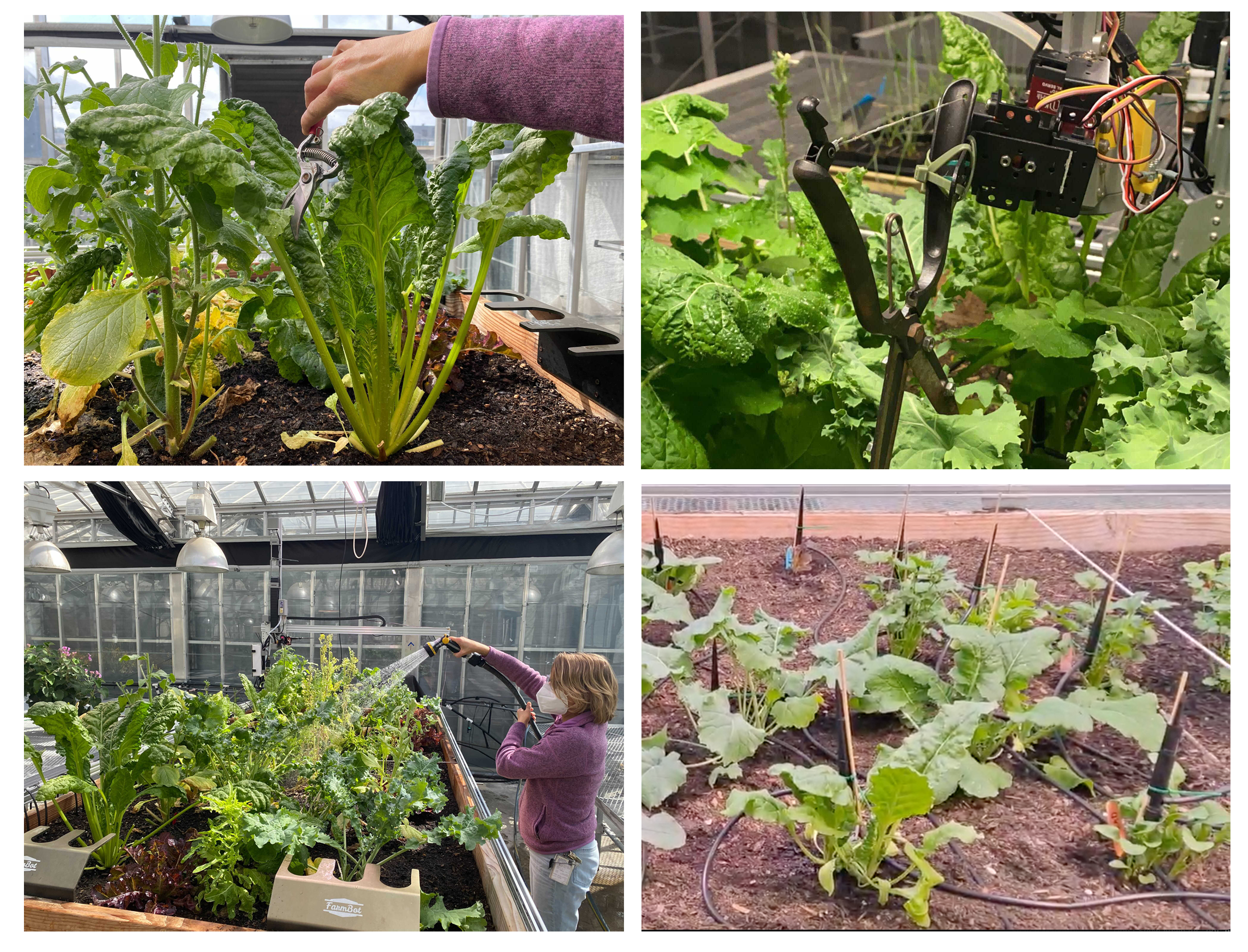}
    \caption{\textbf{Pruning and Irrigation.} \textit{Left:} Human pruning and human irrigation. ~\textit{Right:} Robot pruning and drip irrigation.}

    \vspace{-0.7cm}
\label{fig:setup}
\end{figure}

Each garden has a total of $n$ plants, placed within a planter bed of size $(w, h)$, in cm. For each plant $i \in [0, n)$, the plant has its center coordinates $(x_i, y_i)$, current radius $r_i$, both in cm. Each plant also has a corresponding plant type, $p_i$, which dictates the estimated germination time $g_i$, maturation time $m_i$, and maximum radius $R_i$. The lifecycle of each plant $i$ is defined by five stages: germination, vegetative, reproductive, senescence, and death similar to \cite{avigal2020simulating}. Each stage corresponds to a target volumetric water content (VWC). A garden state on day $t$ includes all information described above for every plant $i \in [0, n)$ at day $t$. Thus, a garden state is defined as:

$$s(t) = [p_i : ((x_i, y_i), r_i), ... ], i \in [0, n)$$

We define coverage, $c(t)$, as the sum of all plant types canopy coverage, $c(t) = \sum_ic_i(t)$, over total area $w \cdot h$ at day $t$ for each plant type $i$. We define garden diversity as:
\begin{align*}
    &\textbf{v}(t) = [c_{i}(t) \cdot (R/R_i)^2, ...], \forall k\\ &d(t) = H(\textbf{v}(t))
\end{align*}
where $H(\cdot)$ is an entropy function, $\textbf{v}(t)$ is a vector of normalized plant types coverage, and $R$ is the average maximum radius over all plant types. Multiplying $c_{i}(t)$ by $(R/R_i)^2$ normalizes each plant type's canopy coverage. We normalize because smaller, less dominant species are less likely to have the same coverage as much larger, faster-growing species. We aim to minimize water consumption while maximizing $c(t)$ and $d(t)$ through irrigation and pruning actions. 
Real experiments are carried out in a raised planter bed of size $3.0$m $\times$ $1.5$ m, located in a UC Berkeley greenhouse. The garden bed is split into two halves with mirrored seed placements generated for the halves as shown in Figure \ref{fig:cycle_at_60}. The left half is tended by horticultural staff with 10+ years of experience and the right half is maintained by AlphaGarden.

\section{Staggered Planting}
In a  polyculture farm setting, plants can grow at different rates, some faster and some slower. To ensure faster-growing plants do not initially out-compete slower-growing ones, we introduce staggered planting, or planting in stages.

Staggered planting allows us to plant slower-growing plants earlier in the planting cycle while faster-growing plants are planted later in the cycle. We adapt staggered planting in the real world to account for variability when plants fail to germinate by transplanting a similar seedling in its place.

\subsection{Staggered Planting in Simulation}
\begin{figure}
    \vspace{0.2cm}
    \centering
    \frame{\includegraphics[width=85mm]{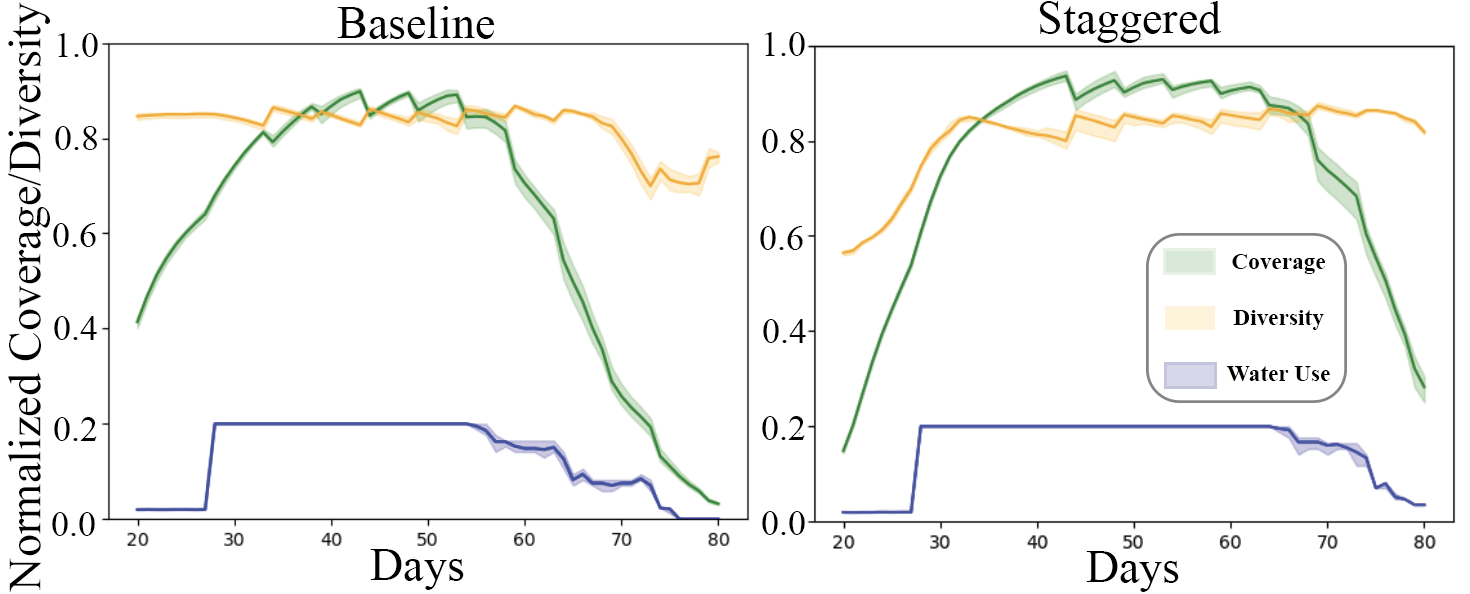}}
    \caption{\textbf{Coverage and Diversity simulation results.} We compare the results of \textit{(left)} normal and \textit{(right)} staggered planting in simulation repeated 5 times. We find higher and sustained coverage in staggered planting through days 30-60. }
    \vspace{-0.6cm}
\label{fig:cov_div}
\end{figure}

    Given an initial seed placement, AlphaGardenSim models the interplant relationships over the course of a garden cycle. We use these dynamics and tuned growth models to predict the diversity and coverage of a garden and compare the results of the two garden cycles. One implements staggered planting and the other does not. The results of 5 randomized trials are shown in Figure~\ref{fig:cov_div}. At day 50, the average normalized canopy coverage is 0.71 and 0.90 respectively for the normal and staggered simulations. Both methods had normalized diversity of 0.86.

\subsection{Staggered Planting in Real}
    To evaluate the results of staggered planting, we compare four 60-day gardens. In the first garden cycle, all plants are seeded on day 1, while in the other garden cycle, the fast growing plants are seeded after 10 days. In the cycle with staggered planting, this allows for the slower-growing plants to get a head start and reach their potential, whereas previously they may have been outcompeted by the faster-growing plants.
    In real-world experiments, staggered planting can also reduce overall water usage, as the larger plants only need to be watered for 50 days.

\section{Variable Irrigation}
\subsection{Modified Analytic Policy}
The analytic policy used in AlphaGardenSim as presented in \cite{avigallearning} decides on policy actions (watering and pruning)  for each plant at every timestep $t$ (in days) and is based on global variables: soil moisture grid, garden diversity, and plant health. For irrigation, the policy decides on a binary amount of water (0 mL or 200 mL) and in simulation, this is done for every plant, every day. The AlphaGarden also uses a planting mix, which has different hydraulic properties than typical field soil. 
Thus, for $N$ plants over $L$ days, 
\begin{align*}
    &\textbf{Max water usage} \approx N \cdot (200 \text{ mL}) \cdot L
\end{align*}

Instead of the default irrigation type where all plants receive either the same volume or no water at all, variable irrigation allows the amount of water received by a plant to be catered to the plant's water demand, growth stage, water uptake, and growth. This potentially allows the overall reduction of water usage.

\begin{figure}
    \vspace{0.2cm}
    \centering
    \frame{\includegraphics[ width=82mm]{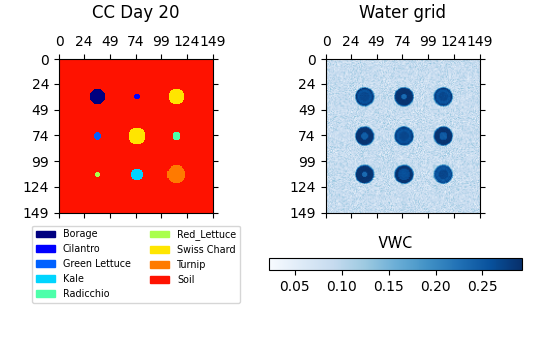}}
    \frame{\includegraphics[ width=82mm]{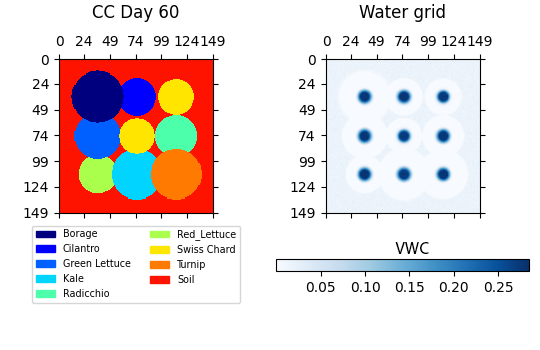}}
    \caption{\textbf{Variable Irrigation Simulation Experiment Overview.} In AlphaGardenSim, we set up experiments to compare  the binary/default implementation of the Analytic Policy with this variable implementation of Analytic Policy. The experiment setting is a  150 cm x 150 cm garden bed, 9 plants placed uniformly and a 100 day cycle.}
    \vspace{-0.6cm}
\label{fig:var_irr_overview}
\end{figure}

\begin{figure*}[t!]
    \vspace{0.2cm}
    \centering
    \frame{\includegraphics[width=160mm]{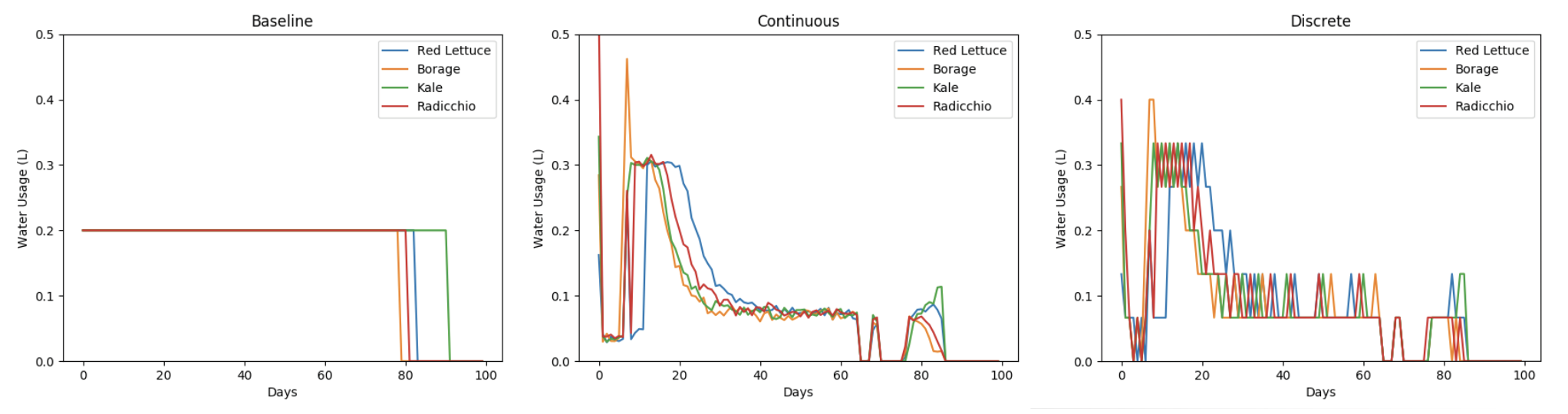}} 
    \caption{\textbf{Variable Irrigation Simulation Experiment Results.} In AlphaGardenSim, we compare three irrigation techniques. For each method we compare the average diversity and coverage across day 20 to 70. \textit{Left} The baseline irrigation method waters 0.2L to each plant everyday, using a total of 272.4 L of water and reaching a coverage and diversity of 0.60 and 0.95. \textit{Center} Continuous Variable Irrigation only uses 143.1 L of water and has a coverage and diversity of 0.58 and 0.95. \textit{Right} Discrete Variable Irrigation waters in increments of [0, 66, 132, 200, 266, 332, 400] mL and achieves a coverage and diversity of 0.58 and 0.95, using 143.6L of water.}
    \vspace{-0.5cm}
\label{fig:var_irr_results}
\end{figure*}

We modify the analytic policy in AlphaGardenSim to better optimize water usage through variable irrigation. For each plant life stage, we assign a desired water amount. These amounts can be roughly estimated from literature and tuned. We assign values as shown in Table~\ref{table:irrigation_vwc}. We compare the default (binary) implementation of the analytic policy with the variable implementation of it and evaluate water usage. This is done on a simulated 150 cm x 150 cm garden bed with 9 plants placed uniformly in a 100-day cycle. An overview of this is shown in Figure~\ref{fig:var_irr_overview}. We create two versions of the variable irrigation policy. The first version is allowed to decide any water amount from 0 mL upwards (continuous). The second version chooses from a fixed list of irrigation values (discrete). We model our real world setup with 16 plants in the simulation. Our results show that variable irrigation reduces mean water usage per day and total water usage by over 47\% in the continuous case. Similarly, in the discrete case, compared with the baseline analytic irrigation policy, mean water usage per day and total water usage reduces over 47\%. Results are shown in Figure~\ref{fig:var_irr_results}

Shrubbler drip emitters can be used to vary the water supply to a zone of the garden or to individual plants. Shrubbler drip emitters allow the adjustment of how much water each plant receives by varying the number of turns on the emitter head and for how long water flows through the emitter. More turns are equivalent to a higher flow rate and running for a longer time means each zone or plant receives more water. In moving from sim to real, we take advantage of this unique property of Shrubbler drip emitters and scale the water requirement for each plant from our experiments in simulation to ensure plant growth. 
\begin{table}
\centering
\vspace{0.15cm}
\scalebox{0.85}{
\begin{tabular}{|c|c|}
\hline
\textbf{Life Cycle Stage} & \textbf{VWC}\\
\hline
Germination & 0.2\\
\hline
Vegetative (growth) &  0.2  \\
\hline
Reproductive (stagnant radii) & 0.3 \\
\hline
Senescence (decaying radii) & 0.2 \\
\hline
Death & 0.1 \\
\hline
\end{tabular}}
\caption{\textbf{Assigned Irrigation VWC} (Volumetric Water Content) for the 5 stages of the life cycle in AlphaGardenSim based on maximum possible VWC found in \cite{avigallearning}.}
\squeeze
\vspace{0.2cm}
\label{table:irrigation_vwc}
\end{table}

\subsection{Closed-Loop Irrigation}
The goal of variable irrigation is to maintain a range of soil moisture levels required for optimal plant growth. However, this assumes consistent environmental variables such as amount of sunlight, temperature, and humidity. To account for this, we introduce closed-loop irrigation tuned by real-world data collected from a previous cycle. By contrast, in open-loop irrigation, the garden space is watered without any feedback from the garden itself. 

The closed-loop system takes input from six TEROS-10 soil moisture sensors \cite{teros10} which automatically upload soil moisture, measured in terms of volumetric water content (VWC), humidity, and temperature data to a cloud storage every thirty minutes. A server queries these sensors periodically and determines whether or not irrigation is required. If required, the server sends information to the Farmduino system on the Farmbot to turn on irrigation. We connect a solenoid valve to the Farmduino and to the drip emitters.

\section{Physical Experiments}

\begin{figure*}[t!]
    \vspace{0.2cm}
    \centering
    \includegraphics[width=160mm]{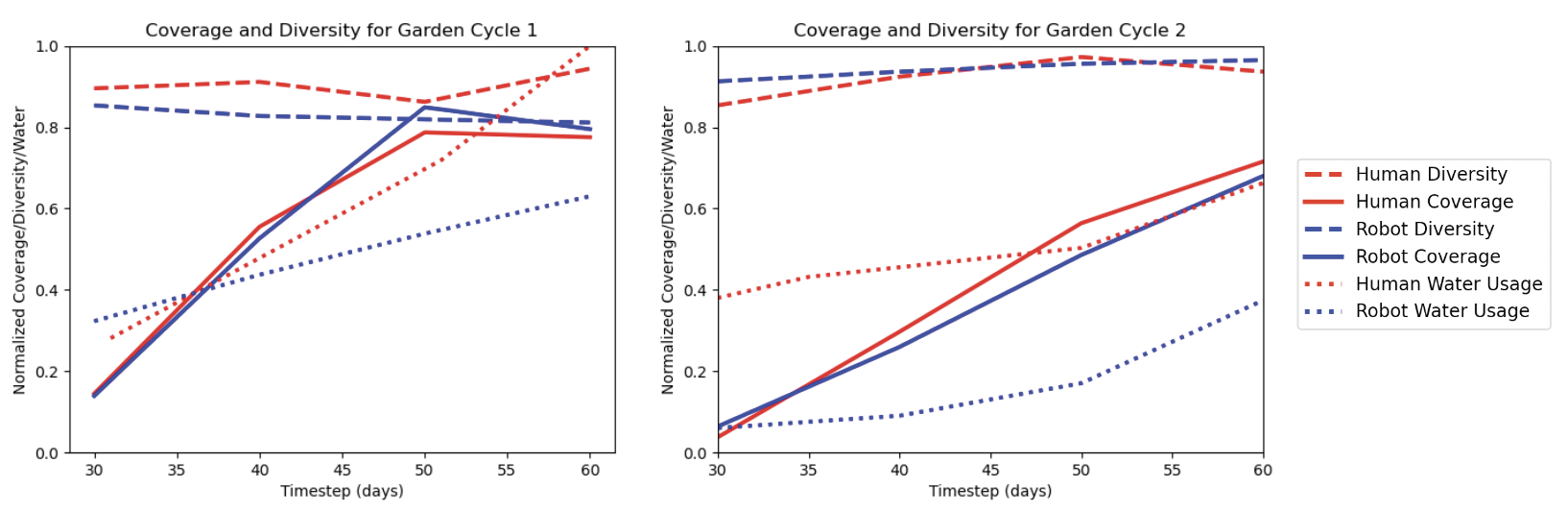}
    \caption{
    \textbf{Performance of Four Physical Gardens.} The graphs show the diversity, coverage, and water usage of four 60 day gardens with the same initial seed placement. We compute metrics starting at day 30. In each garden cycle, one side of the plant bed was tended by the robot and the other by expert gardeners. Irrigation is normalized to be between $[0, 1]$ by dividing by maximum total water usage of $413.5 L$.}
    
    \vspace{-0.2cm}
\label{fig:cycle_1_end}
\end{figure*}
\begin{table*}
\centering
\begin{tabularx}{\textwidth}{|X|X|c|c|X|X|X|} %

\hline

 \multirow{2}{*}{} & \multicolumn{1}{l|}{\textbf{Germination}} & \multicolumn{1}{l|}{\textbf{Irrigation}} & \multicolumn{1}{l|}{\textbf{Pruning}} & \multicolumn{1}{l|}{\textbf{Coverage}} & \multicolumn{1}{l|}{\textbf{Diversity}} & \multicolumn{1}{l|}{\textbf{Water Use(l)}}\\
\cline{2-7}
  & \multicolumn{6}{c|}{\textbf{Cycle 1}} \\
\hline
 Human & \multirow{2}{*}{21 locations} & Every 1-2 days, at their discretion & At their discretion & 0.78 & 0.94 & 413.5 \\

\cline{1-1}
\cline{3-7}
 Robot & &\begin{tabular}[c]{@{}l@{}}Open-Loop at 7 am daily for 60s\\ 3.8L/day (Germination, Mature Growth)\\ 5.33L/day (Early Growth)\end{tabular} &\begin{tabular}[c]{@{}l@{}}Started Day 47\\ 4 pruning sessions\\ 9 plants selected, pruned\end{tabular} & 0.81 & 0.80 & 260.6 \\
\hline
   & \multicolumn{6}{c|}{\textbf{Cycle 2}}\\
\hline
 Human & \multirow{2}{*}{24 locations} & Every 1-2 days, at their discretion & At their discretion & 0.72 & 0.94 & $>=$ 274.1\\
\cline{1-1}
\cline{3-7}
Robot & &\begin{tabular}[c]{@{}l@{}}45 Days:Open-Loop at 7 am daily for 60s\\ 5 Days: Closed-Loop Irrigation \\ 10 Days: Closed-Loop Irrigation \end{tabular} &\begin{tabular}[c]{@{}l@{}}Started Day 30\\ 6 pruning sessions\\ 45 plants selected, 30 pruned\end{tabular} & 0.67 & 0.97 & 154.8 \\
\hline
\end{tabularx}
\caption{\textbf{Physical Experiments Summary:} Germination, Irrigation, Pruning, Coverage, Diversity and Water Use for the human and robot side for cycles 1 and 2}
\label{table:cycle_summary}
\end{table*}

\subsection{Experimental Setup}
We ran two 60-day cycles. Cycle 1 ran from April 15 to June 14, and cycle 2 ran from July 2 to  August 31. For each garden space of 1.5m x 1.5m, we plant sixteen plants (two each for eight types): kale, borage, swiss chard, turnip, radicchio, green lettuce, cilantro, and red lettuce, using the same seed placement arrangement mirrored. Thus there are thirty-two plants in total. After germination, there is the opportunity to thin, transplant, or replace plants depending on germination success. 

The human side is watered at the discretion of the horticulturalists using a nozzle hose fitted with a water flow meter and pruned based on their knowledge. Water usage, observations, and pruning actions are logged. The robot side is watered using drip emitters of the Shrubbler type installed above each plant. As discussed in Section V, Shrubbler emitters can be adjusted to supply different amounts of water based on the turns on their heads. We tested the drip emitters to measure their flow rates, zones of influence, and VWC gain/loss as measured by the TEROS-10 sensors. 
We adjusted the emitter settings accordingly allocating one drip emitter to every plant on the robot side. Specifics are described under each cycle below. 

A water flow meter is also attached to the pipe supplying water to the drip emitters. Water added to both sides of the garden during germination, transplanting, or thinning of the plants is excluded from the total water usage comparison. 

Pruning starts anytime after Day 30 and runs till the end of the cycle. Peak growth periods are from Days 40 - 50.

To analyze the garden state, we label each plant type and compute diversity and coverage metrics every 10 days. 
Figure~\ref{fig:setup} shows an expert horticulturalist pruning and watering one side of the garden and the emitters irrigating the robot side.

\subsection{Results}

\subsubsection{Human Intervention}
Our test compares the pruning and irrigation decisions of AlphaGarden versus humans. However, on the robot side, at times, human intervention was required in the form of manually cutting prune points, adjusting selected prune points, and adjusting the orientation and depth of the pruning tool as needed. 

The `human' side of the experiment was tended by UC Berkeley Oxford Tract Greenhouse staff who have on average, 10 years of professional horticultural experience.

\subsubsection{Cycle 1}
\paragraph{Germination}
Of the thirty-two plants locations, no radicchio, green lettuce, and red lettuce germinated on either side. Green lettuce and red lettuce were replaced from local nurseries with seedlings of the same plant types in an early growth stage. Radicchio was difficult to find and was replaced with arugula seedlings, which have a similar growth pattern. One turnip plant germinated on the robot side but none germinated on the human side. Turnip was replaced with mustard green seedlings on both sides of the garden. Cilantro germinated on the human side but not on the robot side. Two excess cilantro sprouts were moved from the human side to the robot side. All other plants germinated. However, one swiss chard on the robot side germinated but was stunted throughout the entire cycle. It was later found out that some of the seeds used in this cycle had been in storage for more than a year and this may have affected their viability. No additional thinning was carried out. 
\paragraph{Irrigation}
On the human side, the garden was watered roughly every 1-2 days. On the robot side, open-loop irrigation took place at 7:00 AM daily for 60 seconds. The entire cycle was divided into three periods depending on the plants' estimated lifecycle: germination (16 days), early growth (19 days), and mature growth (25 days). Plants are divided into two groups based on their perceived water needs. Group 1 consisted of kale, borage, swiss chard, and turnip while group 2 consisted of radicchio, green lettuce, cilantro, and red lettuce. During germination, the emitters above group 1 plants were set to seven (7) turns which gave about 284 mL while the emitters above group 2 plants were set to six (6) turns which gave about 191 mL totaling about 3.8L per day. During early growth, the emitters above group 1 plants were set to eight (8) turns which gave about 383 mL while the emitters above group 2 plants were set to seven (7) turns which gave about 284 mL totaling about 5.33L per day. During mature growth, the emitters were returned to the same setting as in germination. 

\paragraph{Pruning}
Automated pruning actions on the robot side started on day 47 and were executed with an interval of 3 days between pruning sessions. AlphaGardenSim autonomously determines what plants and what leaves to prune, utilizing the pruning pipeline as described in \cite{presten_automated_2022}. Pruning actions were then executed by a human and/or the Farmbot. There were a total of four pruning sessions, nine plants were selected by AlphaGardenSim and were pruned.

\paragraph{Coverage \& Diversity}
We found final coverage and diversity of $0.81$ and $0.80$, respectively, while using $260.6$ liters of water on the robot side while the human side attained a coverage of $0.78$ and a diversity of $0.94$ and water usage of $413.5$ liters. Figure~\ref{fig:cycle_1_end} shows this. AlphaGarden achieved comparable diversity and coverage with 37\% less water.

\subsubsection{Cycle 2}
\paragraph{Germination}
In cycle 2, we implemented staggered planting. Four plant types known to be slower-growing, cilantro, green lettuce, red lettuce, and radicchio, were planted from Day 1. Four plant types known to be faster-growing, kale, borage, swiss chard, and turnip, were planted from Day 11. Of the thirty-two plants locations, plants germinated in twenty-four locations.  Similar to cycle 1, we transplanted and purchased plants for the locations without germination. 
As in Cycle 1, it was difficult to find radicchio seedlings of comparable growth and so it was replaced completely with arugula which has similar growth tendencies. All other plants including all of the faster-growing plants germinated.  Manual thinning was carried out on Day 28 leaving all locations with 1-3 plants each. 

\paragraph{Irrigation}
The irrigation schedule on the human side was similar to Cycle 1 with the garden watered roughly every 1-2 days. However, on the robot side, a number of changes were made. First, the same drip emitter turn settings were used throughout the plants' lifecycle. Group 1 plants, i.e. kale, borage, swiss chard, and turnip had their emitters set to seven (7) turns while group 2 plants i.e. radicchio, green lettuce, cilantro, and red lettuce had their emitters set to six (6) turns. The emphasis was to regulate how much water the garden received by how long the flow was.  For 45 days in the cycle, open-loop irrigation took place at 7:00 AM daily for 60 seconds. From day 46 to day 60, we tested out closed-loop irrigation with two different threshold and time pairs: from day 46 to day 50, we used a normal and low metric $0.25 $ and $0.18 L / m^2$, watered for 60 seconds for the normal threshold and 120 seconds for the lower threshold and watered once every 24 hours. From day 50 to day 60, we used a normal and low metric of  $0.25$ and $0.21 L / m^2$, respectively, and watered for 60 seconds every 24 hours for the normal threshold and for 343 seconds every six hours for the lower threshold.

In Cycle 2, the flow meter attached to the robot side malfunctioned and did not accurately record water use. Therefore we only have water usage estimates based on the irrigation schedule. Similarly, on the human side, the flow meter was erroneously reset on Day 47 and some water usage data was lost. As a result, we can only provide a lower bound on how much the human side used in Cycle 2. We estimate the human side used at least $274.1$ Liters while the robot side used a total of $154.8$ Liters: $42$ Liters (Day 1-46), $9$ Liters (Day 46-50), and $103.8$ Liters (Day 51-60). 

\paragraph{Pruning}
Automated pruning actions on the robot side were started on day 30 and executed with an average interval of 5 days between pruning sessions. Similarly, AlphaGardenSim autonomously determined what plants and what leaves to prune, utilizing the pruning pipeline as described in \cite{presten_automated_2022}. Pruning actions were then executed by a human and/or the Farmbot. There were a total of six pruning sessions during which 45 plants were selected by AlphaGardenSim and 30 were pruned. The remaining 15 were either missing prune points or there was another error in the pipeline. 

\paragraph{Coverage \& Diversity}
We found final coverage and diversity of $0.67$ and $0.97$ respectively while using $154.8$ liters of water on the robot side while the human side had a coverage of $0.72$ and a diversity of $0.94$ with at least $274.1$ Liters. In this case, AlphaGardenSim achieved comparable diversity and coverage while using at least $44$\% less water.

Table~\ref{table:cycle_summary} presents a summary of the physical experiment cycles.

\section{Limitations}
We studied staggered planting policies based on learned plant growth models which were inspired by traditional practices in polyculture, but can decrease coverage.

Slightly increased exposure to sunlight on the human side due to garden roof angle may have contributed to the results. 

Variations between cycle 1 and cycle 2 may be due to growing season variations.

It is possible to further improve irrigation tuning by extending AlphaGardenSim to work with plant uptake models such as proposed in Faria et al. \cite{Faria}. Models of plant water uptake are necessary to determine optimal irrigation programs for gardens and will allow us together with the soil sensors to customize water for each plant not just to its period in the lifecycle or to how much water is in the soil but a combination of these and how each plant is growing.

\section{Conclusion}
Across two growth cycles, the robot achieved comparable diversity and coverage to human experts while using significantly less water. AlphaGarden has thus passed the Turing Test for gardening. Much remains to be done.

\section*{Acknowledgements}
\vspace{-0.1cm}
\relsize{-1}
This research was performed at the AUTOLAB at UC Berkeley in affiliation with the Berkeley AI Research (BAIR) Lab, and the CITRIS "People and Robots" (CPAR) Initiative. This research was supported in part by Siemens. We thank our colleagues who provided helpful feedback and suggestions, in particular Yahav Avigal, Wisdom Agboh, Muyan Jiang, Kishore Srinivas, and the staff of the UC Berkeley Oxford Tract Greenhouse: Kristina Chan, Ben Russell, Adam Hale, Jillian Demus, and Bradlee Mcanally.

\nocite{goldberg2002beyond, harper1977population, toshioK}
\bibliographystyle{IEEEtran}
\bibliography{IEEEabrv,references}

\end{document}